\documentclass[twocolumn,epsfig,aps]{revtex4}   
\usepackage{epsfig}
\usepackage{graphicx}

\begin{document}
\title{A New Computational Framework For 2D Shape-Enclosing Contours}
\author{B.R.~Schlei\footnote{\textit{Email address:} \texttt{schlei@me.com}}}
\affiliation{
Los Alamos National Laboratory, Theoretical Division, T-1,
P.O. Box 1663, Los Alamos, NM 87545
}

\begin{abstract}
\noindent
In this paper, a new framework for one-dimensional contour extraction 
from discrete two-dimensional data sets is presented. Contour extraction 
is important in many scientific fields such as digital image processing, 
computer vision, pattern recognition, etc. This novel framework includes 
(but is not limited to) algorithms for dilated contour extraction, contour 
displacement, shape skeleton extraction, contour continuation, shape 
feature based contour refinement and contour simplification. 
Many of the new techniques depend strongly on the application of a 
Delaunay tessellation. In order to demonstrate the versatility of this 
novel toolbox approach, the contour extraction techniques presented here
are applied to scientific problems in material science, biology, 
handwritten letter recognition, astronomy and heavy ion physics.\\

\textit{Keywords:} Contour, Isocontour, Edge, Unstructured Grid,
Delaunay tesselation, Skeleton, Shape morphology, Material surface,
Bacterial colony, Handwritten letter recognition, Constellation,
Freeze-out hyper-surface\\
\end{abstract}
\maketitle

\section{Introduction}
In two spatial dimensions, a lower-dimensional interface which partitions
a two-dimensional (2D) space into separate subdomains with nonzero areas
is called a contour. 2D spaces can be either continuous or discrete with
respect to a field quantity, which is defined across that particular space.
For example, a 2D gray-level image represents a discrete 2D space with 
respect to the field quantity gray-level. Its area, which is covered by 
the image, is broken into many regular 2D cells, i.e., pixels ($=$ picture 
elements); each pixel has a constant shade of gray.\\
\indent
Many papers~\cite{REF01}--\cite{REF06} have been written on the extraction 
of one-dimensional (1D) contours from 2D image data. It is not trivial to
define a contour for a discrete space. For example, one has to specify how 
the final contour should be supported. Some contour extraction algorithms
yield contours which only connect the centers of edge pixels 
(\textit{cf.}, e.g., Ref.~\cite{REF07}; an edge pixel
is a pixel which is considered to represent a part of the boundary of a
certain region of interest within a given image). Others may allow for 
the usage of points that lie on the boundary between two pixels 
(\textit{cf.}, e.g., Ref.~\cite{REF08}).
More complications may arise, if a resulting contour is not
closed or if it encloses an area which equals zero (in the latter case,
we call a contour degenerate). Furthermore, a contour may be 
self-intersecting and as a result it may enclose more than one of the 
2D regions.\\
\indent
One has also to consider the level of information, which is provided for
building contours. In some applications, a 2D image is preprocessed 
through image segmentation~\cite{REF08}--\cite{REF10}, i.e., pixels are 
grouped together into so called blobs. Very often, a contour extraction 
method is subsequently applied to the generated image blobs. 
In other applications, a 2D image is preprocessed by an edge detector 
(\textit{cf.}, e.g., Ref.~\cite{REF11}), i.e., edge pixels are identified  
which are assumed to describe the transition of two different neighboring 
regions (note, that an edge pixel is a 2D object rather than a section of 
a 1D contour). A contour extraction method may then be applied to the 
resulting edge pixels. In particular, complications may arise, if 
the edge pixels provided by an edge detector form only partially
connected chains, or if the transition region (given by the edge pixels) 
of two zones in a 2D image exceeds the width of more than one pixel.\\
\indent
Ref.s~\cite{REF01}--\cite{REF10} demonstrate that many different image 
processing problems have resulted in many different approaches for 
building 1D contours from 2D image data. It is therefore the intent of 
this paper to provide a single computational framework for building 2D 
shape-enclosing contours from various different types of 2D discrete data
sets. The considered data sets will include both, 2D gray-level images
and 2D simulation grids, e.g., 1+1D (i.e., 1D space + 1D time) 
hydrodynamic simulation data. 
The framework presented here will handle many different 2D image processing
problems with one and the same set of tools. In particular,
this novel framework will provide solutions for the problems described 
above and for others which one may be faced with when extracting contours 
from discrete 2D data sets. Note, that the results of this framework will
depend strongly on the quality of preprocessed data, e.g., the 
segmentation of 2D image data. The latter subject is beyond the scope 
of this paper. We will rather focus on some of the geometrical features of 
the properly preprocessed 2D data.\\
\indent
This paper is structured as follows. In the next section, the contour 
extraction framework is explained. The topics which are covered in this 
section include (but are not limited to) dilated contour 
extraction~\cite{REF12}, contour displacement, shape skeleton extraction, 
gap closure or contour continuation, shape feature based contour 
refinement and contour simplification. This section is followed by an 
application section, where this novel toolbox approach for 1D contour 
extraction is applied to scientific problems in material science, biology,
handwritten letter recognition, astronomy and heavy ion physics.
A summary will conclude this paper.

\section{The Contour Extraction Framework}

In this section, we decribe how to extract 1D contours from discrete 2D
spaces. As mentioned above, a discrete 2D space could be represented by a 
2D image. 
However, a proper discrete 2D space could also be given through the union 
of all triangles resulting from a 2D Delaunay tessellation~\cite{REF13} 
(including some additional field quantities that characterize each 
triangle further), etc.~\cite{REF02}. 
In the next section however, we shall restrict ourselves - without loss 
of generality - to the case of 2D images. Note, that the following contour 
extraction algorithm~\cite{REF12}, which is also known under the
name DICONEX, has been implemented efficiently into software~\cite{REF14}.

\subsection{DICONEX - DIlated CONtour EXtraction}

The DICONEX algorithm~\cite{REF15}--\cite{REF17} always yields perfect 
contours for both, binary and gray-level images. The contours are perfect 
in the sense that they are non-selfintersecting and non-degenerate, i.e., 
the contours always enclose an area larger than zero.
DICONEX operates on a segmented 2D image (or comparable data structure).  
Fig.~1.a shows a binary image with 25 white and 11 gray pixels.\\
\indent
As a first step, a set of disconnected vectors is constructed 
which separates white pixels from gray ones. Each vector is attached to 
a pixel with its origin and its endpoint in such a way that the 
pixel always lies to the left of the vector (\textit{cf.}, Fig.~1.b). 
This ensures the counterclockwise circumscription of all pixels (or 
clusters of pixels) by the vectors. Conversely, all holes in a 
pixel cluster (blob) are circumscribed clockwise (\textit{cf.}, 
Fig.~1.c and Fig.~1.d). Note, that in order to accomplish this pixel 
enclosure by oriented vectors, it is only necessary to consider the four 
nearest neighbors of any given pixel, i.e., its upper, left, lower, 
and right pixel neighbor (\textit{cf.}, Fig.~1.b). Each vector is 
unique, double counting can never occur. Furthermore, there is no 
specific order required in which the neighborhood of any given pixel 
is evaluated. Therefore, this processing step is totally parallel.\\
\begin{figure}[t]
\epsfig{width=8.5cm,figure=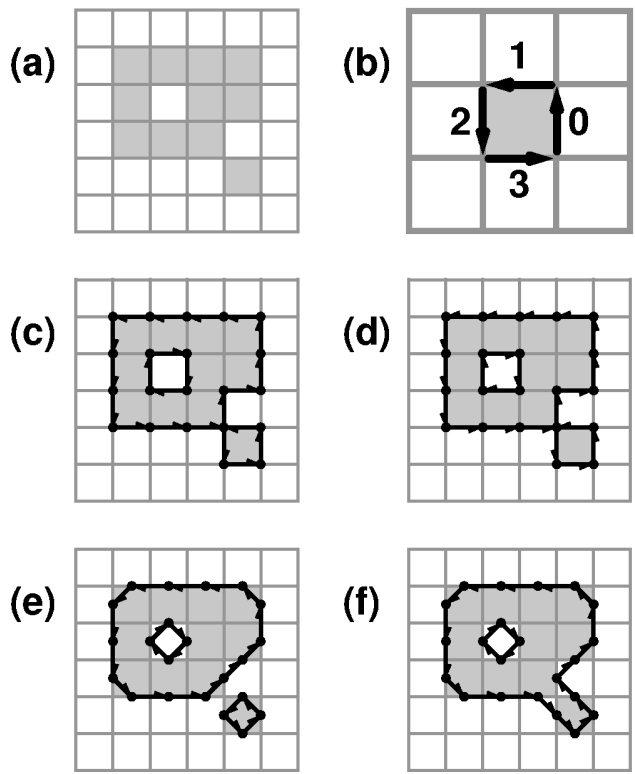}
\vspace{-0.1in} \caption{
(a) Initial binary image; (b) contour vectors for
a single pixel; (c) left-turning and (d) right-turning contour vectors 
for initial binary image; (e) dilated contours for left-turning contour 
vectors (``disconnect'' mode); (f) dilated contours for right-turning 
contour vectors (``connect'' mode).
} \label{multib_01}
\end{figure}

\indent
In the second step, which is linear, connected loops are 
constructed from the previously generated contour vector set. Each vector
is attached with its origin to the end point of another vector. When
connecting the vectors to contours, one starts with a single vector.
Note, that each vector will contribute to the set of contructed contours 
only once.
If one attaches the next vector, which has its origin attached to the end
point of the current vector, one is always faced with either one of the two
following options. Either there is one vector or there are more than
one vectors (\textit{e.g.}, two in the case of pixel processing) connected 
to a single vector.
In the latter case, the user has to decide, if a pixel should be 
disconnected or connected to the current blob whose enclosing 
contour is being constructed; we either choose always a left-turn 
(\textit{cf.}, Fig.~1.c) or always a right-turn (\textit{cf.}, Fig.~1.d,)
when connecting the vectors. 
In doing so, we ensure a consistent choice for building the contours
while tracing the sequence of vector origins. Additionally, we either 
weaken the connectivity between pixels which touch each other only in one 
point or we strengthen it. In fact, the dilated versions of the contours 
will lead either to a total separation (\textit{cf.}, Fig.~1.e) or to a 
merging (\textit{cf.}, Fig.~1.f) between two pixels that share only one 
common (pixel corner) point.
Fig.~1.c and 1.d show the left-turn and right-turn contours, respectively, 
for the rendered bi-level (binary) image shown underneath.\\
\indent
Finally, the vector origins are replaced with the midpoints between the 
origins and the endpoints of the contour vectors without changing the 
connectivity among the vectors. As a result, one obtains modified contours 
which represent a dilation of the centers of the blob edge pixels. Figs.~1.e
and 1.f depict the dilated contours according to the technique outlined 
here. Note, that in some of the following figures the arrow heads of the 
vectors will be omitted.

\begin{figure}[b]
\epsfig{width=8.5cm,figure=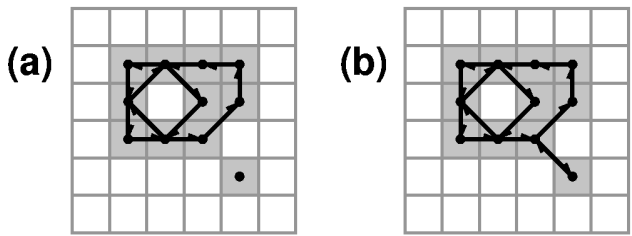}
\vspace{-0.1in} \caption{
(a) boundary pixel tracing contours (``disconnect'' 
mode); (b) boundary pixel tracing contours (``connect'' mode).
} \label{multib_02}
\end{figure}

\subsection{Boundary pixel tracing contours}

Another type of contour can be obtained by tracing the pixel boundary 
of image blobs~\cite{REF07} while connecting the pixel centers. 
Such boundary pixel tracing 
contours (BPTCs) have the advantage that they can be obtained with very
little memory requirement while making use of so called chain codes. 
A chain code 
is a sequence of directions, typically indicating the shortest path to 
one of the next eight neighbors of a given pixel. However, if holes 
are present within the shapes of a given pixel cluster, an algorithm may be 
unable to construct the contours successfully~\cite{REF06}.\\
\indent
The DICONEX algorithm can also be used to construct BPTCs. 
Instead of finally replacing the origin of a contour vector with the 
midpoint between its origin and its endpoint, the origin of a contour 
vector is moved to the center of the pixel to which the contour vector 
is attached to initially. Fig.~2 shows BPTCs according to this technique.\\ 
\indent
Note, that there are two contour solutions possible for the pixel 
configuration shown in Fig.~2, because a pixel which is in contact with 
another one in only one point may be disconnected from or connected to its 
apparent partner. In fact, it is this ambiguity that may cause an algorithm 
to crash in its effort to construct BPTCs successfully, because it may not 
have accounted for such cases consistently. We would like to stress that 
the BPTCs may be self-intersecting and - partially or fully - 
degenerate (\textit{cf.}, e.g., the point in Fig.~2.a is a fully
degenerate contour).

\subsection{Isocontours}

Sometimes, 2D gray-level image data are processed with the intent to 
extract isocontours. An isocontour is a contour which has a constant 
value at all of its supporting points with respect to the field 
quantity that has been used for the contour extraction. Because the 
points which support a DICONEX contour always coincide with the midpoints 
of the boundary edge between two pixels, dilated contours are in
general not isocontours. However, these contours can be transformed
into isocontours very easily.\\
\indent
Fig.~3.a shows a 2D gray-level image with 36 pixels. Here, black pixels 
have a gray-level of value zero, whereas white pixels have a gray-level 
of value 255. We shall now construct an isocontour corresponding to a
gray-level of value 100. First, all pixels with a gray-level of value
larger than or equal to 100 are enclosed with contour vectors as depicted 
in Fig.~3.b. 
From these contour vectors a dilated contour is constructed as shown in 
Fig.~3.c. In Fig.~3.c, additional vectors are drawn for each of the points 
that support the DICONEX contour. We will refer to these additional 
vectors as range vectors. Each range vector connects the centers of the 
pair of pixels that share the boundary edge of the initial contour vectors. 
The origin of a range vector coincides with the pixel center of the 
larger gray-level value, whereas its endpoint coincides with the pixel 
center of the smaller gray-level value.\\
\begin{figure}[t]
\epsfig{width=8.5cm,figure=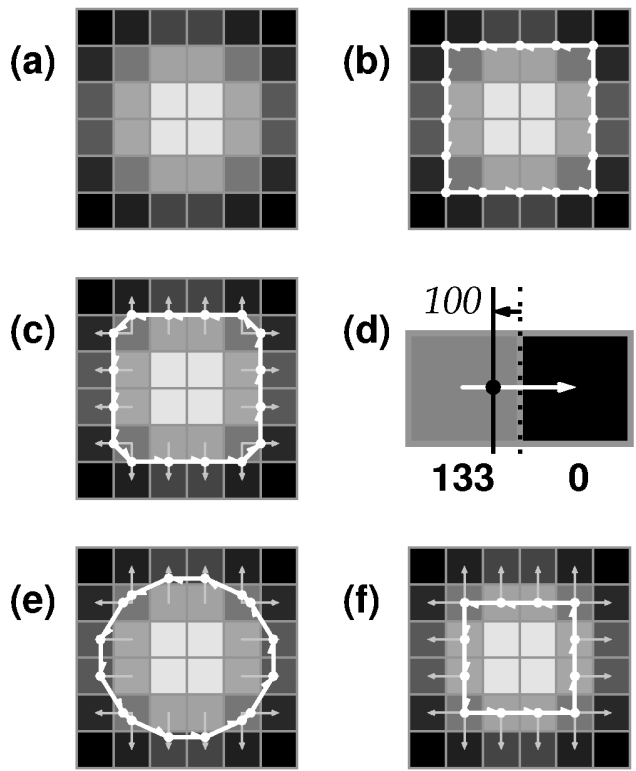}
\vspace{-0.1in} \caption{
(a) Initial gray-level image; (b) as in (a), but 
with left-turning contour vectors; (c)  as in (a), but with range 
vectors and dilated contour; (d) contour displacement for a pixel 
pair (see text); (e) gray-level image superimposed with range vectors 
and an isocontour of value 100; (f) as in (e), but with boundary pixel 
tracing contour.
} \label{multib_03}
\end{figure}
\indent
Fig.~3.d illustrates how DICONEX contours can be
transformed into isocontours. Two pixels - one with a gray-level
value of 133, the other one with a zero valued gray-level - are initially
separated by a DICONEX contour section that is located exactly in
the middle between them (dotted line). A range vector connects the
centers of the two pixels. It defines the bounds within which the support 
point for a dilated contour may be displaced. The centers of each pixel 
are assumed to correspond exactly with their gray-level values. 
The location of the support point for the isocontour is located closer to 
the center of the pixel with the gray-level of value 133. 
Note, that within this paper we use linear interpolation of the 
gray-levels. In Fig.~3.e, the dilated contour of Fig.~3.c has been 
transformed into an isocontour representing a gray-level of value 100.\\
\indent
While using range vectors, one may transform a dilated contour also into a 
boundary pixel tracing contour (\textit{cf.}, the previous section). 
One simply has to move all points which support a dilated contour to the 
origins of the corresponding range vectors as shown in Fig.~3.f.\\
\indent
Note, that the decision whether contours should be left- or right-turning
can also be made locally when using gray-level images. Then one could
consistenly use left-turns, if the turning point under consideration has 
an interpolated gray-level value above the isocontours gray-level value.
Right-turns would then be taken otherwise, or vice versa.

\begin{figure}[b]
\epsfig{width=8.5cm,figure=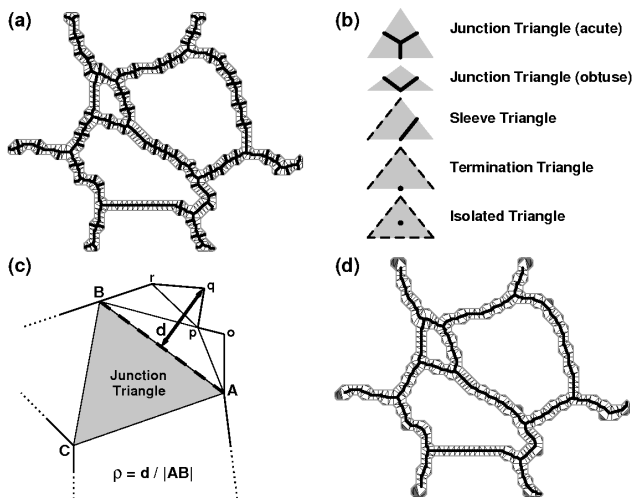}
\vspace{-0.1in} \caption{
(a) A shape with internal triangle decomposition and
skeleton; (b) triangles with shape skeleton segments (the dashed lines 
indicate segments of a shape contour); (c) geometric pruning for 
skeletons (see text); (d) as in (a), but with a pruned skeleton.
} \label{multib_04}
\end{figure}

\subsection{Delaunay tessellation and shape skeleton}

Edge detection algorithms such as the Canny edge detector~\cite{REF11}
return when applied to a 2D gray-level image a set of 2D pixels rather 
than 1D contours or contour segments. Here, we prefer to generate 1D 
contours from a given set of edge pixels. In the following, a set of edge 
pixels will represent 2D shapes from which we are going to extract their 
skeletons in terms of 1D contours or at least contour segments.\\
\indent
First, we generate DICONEX contours for the edge pixel blobs. Next, the 
dilated contours and their supporting point set will be processed with a 
constrained Delaunay tessellation (CDT). Note, that no additional 
``Steiner'' points~\cite{REF13} will be added to this tesselation. 
In Fig.~4.a, the interior of a 2D shape is decomposed into a set of 
triangles after application of a CDT. This is the key step that allows 
for the skeleton extraction of the shape~\cite{REF18}--\cite{REF20}.
The triangles originating from the Delaunay tessellation can be 
classified into four types, namely those with three, two, one or none 
external (i.e., polygonal boundary) edges.
Accordingly, we denote these triangles as ``isolated,'' ``terminated'' 
(T), ``sleeve'' (S), and ``junction'' (J) triangles, respectively.
For each triangle, line segments or single points can be drawn 
(\textit{cf.}, Fig.~4.b), which in their union represent a skeleton of 
the processed shape.\\
\indent
Fig.~4.a shows in addition to the CDT of the 2D shape's interior also its
shape skeleton. Because of small variations along the shape's enclosing 
contour, structurally unimportant skeleton features may occur. However,
these can be removed according to the following pruning 
method~\cite{REF19}. 
Initially, all junction triangles are evaluated based on their nesting 
level within a given shape. Junction triangles which are
nested most deeply are processed first, those which are closest to the
shapes contour(s) are processed last. 
Fig.~4.c provides an example for
pruning. For a given junction triangle $ABC$, we consider its 
edge $AB$ and the shapes contour section $AopqrB$. 
For each point of the set $P = \{o,p,q,r\}$, we compute the distance 
$d$ to the junction triangle's edge $AB$. Let $\rho \equiv d / |AB|$ be
the ratio of morphological significance. If for any of the points in $P$, 
the ratio $\rho$ exceeds or equals a fixed threshold, $\rho_0$, 
the contour section and the corresponding skeleton branch will be 
preserved, otherwise closed polygon $AopqrBA$ will be removed. 
As a consequence, the junction triangle may turn into a sleeve triangle, 
where the triangle edge $AB$ becomes a new (virtual) shape boundary edge.\\
\indent
Note, that before the junction triangle is turned into a sleeve triangle, 
the above algorithm is also applied to the edges $BC$ and $CA$. Hence, a 
junction triangle could even turn into a terminal or into an isolated 
triangle. In Fig.~4.d, the pruned skeleton is shown for the inital 2D 
shape of Fig.~4.a with a threshold $\rho_0 = 0.6$.  

\subsection{Frame addition and gap closure}

\begin{figure}[t]
\epsfig{width=8.5cm,figure=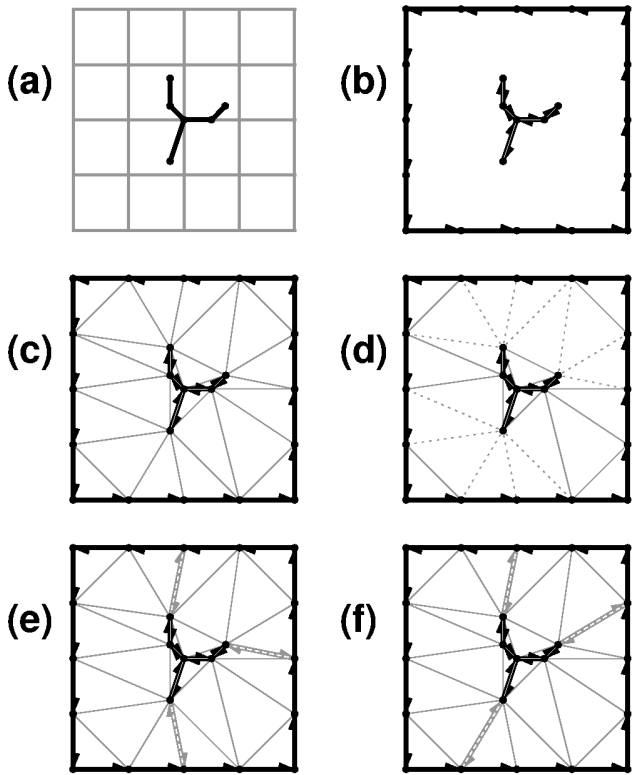}
\vspace{-0.1in} \caption{
(a) A pixel frame superimposed with a shape skeleton;
(b) shape skeleton vectors and frame vectors; (c) as in (b), but with 
additional Delaunay triangular mesh; (d) as in (c), but with suggested 
gap closure lines (dotted); (e) as in (c), but with gap closing vector 
pairs (gray) of shortest length; (f) as in (c), but with vector pairs 
(gray), which attempt to conserve directions.
} \label{multib_05}
\end{figure}

Edge pixels are very often generated in order to decompose 2D image data 
into several disjunct regions. However, their shape skeletons may not 
always provide a complete partitioning of the underlying 2D space. In the
following, we apply a frame addition and gap closure technique~\cite{REF17}
to an initially given skeleton in order to partition an image into several
contour-enclosed non-zero areas.\\
\indent
In Fig.~5.a, an image area of 4x4 pixels is shown, which is superimposed
by a shape skeleton. First, two opposite vectors are assigned to each line 
segment of the shape skeleton, so that the length and orientation of each 
vector pair coincides with the length and orientation of each of the 
skeleton's line segments (\textit{cf.}, Fig.~5.b). 
Next, a frame of vectors is added around the original image given in Fig.~5.a. 
These outer frame vectors are arranged counterclockwise around the original 
image as shown in Fig.~5.b. 
The point set which supports the frame's vector set is sampled at the rate of 
pixels available along a side of the image. In particular, it includes the 
four corners of the image.\\
\indent
In order to close the gaps between the skeleton and the frame, a CDT is 
applied to the point set, which supports the skeleton and the image 
frame (\textit{cf.}, Fig.~5.c). 
Certain edges (shown dotted in Fig.~5.d) of the unstructured grid connect 
the terminal points of the skeletons' limb-like arcs with the outer image 
frame. 
Now, one could either select the shortest edges (\textit{cf.}, Fig.~5.e) or 
the edges which preserve mostly the orientation of the terminating vector 
pair in a skeletons' limb (\textit{cf.}, Fig.~5.f) among the edges 
bridging a gap between a terminal point and a frame point. 
In either event, the gaps will then be closed with vector pairs of opposing 
direction. 
Finally, all vectors can be connected to region-enclosing contours exactly 
the same way as it is prescribed by the DICONEX algorithm (here, only 
left-turns will be performed whenever junctions are encountered). 
Note, that this gap closure technique can be easily extended to general 
gap closure or edge continuation between disjunct skeletons.

\subsection{Contour refinement using shape features}

\begin{figure}[b]
\epsfig{width=8.5cm,figure=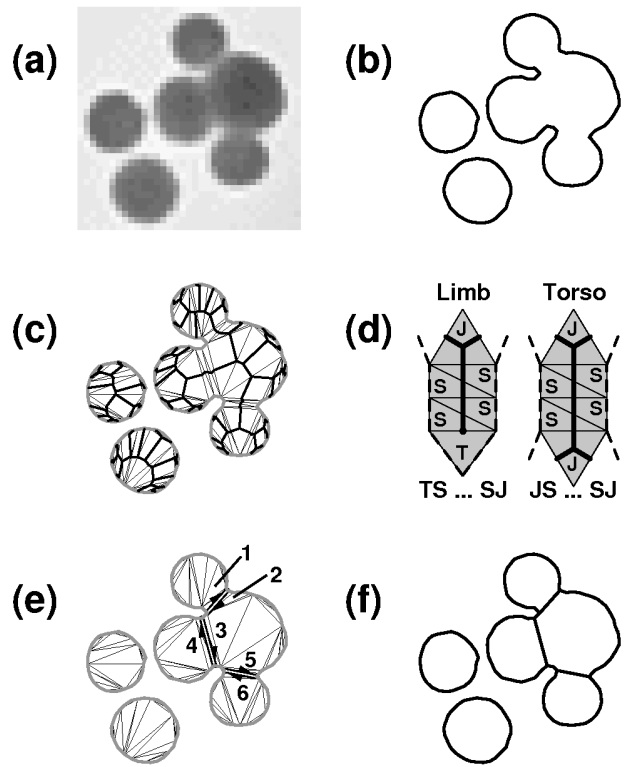}
\vspace{-0.1in} \caption{
(a) Initial gray-level image; (b) three isocontours 
for the dark spots in image (a); (c) shape skeletons, which are enclosed 
by the contours in (b); (d) a limb and a torso (see text); 
(e) torso-splitting vector pairs (black); (f) six final shape-enclosing 
contours.
} \label{multib_06}
\end{figure}

The contours which may be extracted from 2D image data very often do 
not resemble the shapes which a human observer might expect.
For example, Fig.~6.a shows a gray-level image with six circular shaped 
dark pixel clusters, of which four of them slightly overlap. Isocontour
extraction as described previously in this paper results in three 
shape-enclosing contours as shown in Fig.~6.b for an isovalue of 130. 
A human observer might have expected the extraction of six contours instead.
In the following, we describe a technique~\cite{REF21} which allows us to 
increase the initially found number of contours by three while making use 
of higher-level shape features.\\
\indent
In Fig.~6.c, the interiors of the three contours have been
decomposed through a CDT and the unpruned shape skeletons are depicted. 
Ref.s~\cite{REF18}--\cite{REF20} have made clear the value of Delaunay 
triangulations in obtaining structurally meaningful decompositions of 
shapes into simpler components. 2D shapes can be decomposed into generic 
shape components, the so called limbs and torsos~\cite{REF18,REF19}. 
A ``limb'' is a chain complex of pairwise adjacent triangles 
which begins with a junction triangle and ends with a termination triangle. 
A ``torso'' is a chain complex of pairwise adjacent triangles, which both 
begins and ends with a junction triangle (\textit{cf.}, Fig.~6.d). Note, 
that a string-like shape, i.e., $TS...ST$, is a degenerate limb, whereas 
a torus-like shape, i.e., $S_oS_1...S_NS_o$, is a degenerate torso.\\
\indent
Here (\textit{cf.}, Fig.~6.e), the torsos  play an important role which 
are encapsulated by the junction triangle pairs $1\&2$, $3\&4$, and $5\&6$. 
For these torsos only, none of the longest edges of both encapsulating 
junction triangles face their other junction triangle partner. If vector 
pairs are placed with opposing directions at the narrowest local widths 
for these torsos, we may break-up the considered contour into four. 
Note, that the initial shape-enclosing contours consist of vector sequences, 
which enclose the shapes counterclockwise. After insertion of the vector 
pairs, all vectors can be newly connected to region-enclosing contours as it 
is prescribed by the DICONEX algorithm (once again, only left-turns will be 
performed whenever junctions are encountered).
Hence, we end up with four separate contours for the four overlapping 
circular shaped dark pixel clusters, whereas the other two contours remain 
unchanged (\textit{cf.}, Fig.~6.f).\\
\indent
Very often it is possible to reduce the number of points which support 
region-enclosing contours. For example, it is sufficient to represent
contour sections which form straight lines by two points only. 
The same may apply for contour segments which turn only very little. 
For the down-sampling of contours various 
techniques~\cite{REF22}--\cite{REF24} have been proposed. 
In this paper, we shall use a method (\textit{cf.}, Ref.~\cite{REF17}) 
which requires only local contour information when points are evaluated for 
removal.\\
\indent
This section concludes the theoretical part of the here presented framework.

\section{Applications}

In this section, we demonstrate the versatility of the above described 
toolset for contour extraction, etc., by considering a rather diverse group 
of applications.
First, we process an image generated by electron backscattered diffraction 
in order to better quantitatively describe a granular metal surface. 
In a second application, we are going to improve on the counting of 
bacterial colonies within a given image of a Petri dish. 
Thirdly, we process an image showing a handwritten Japanese letter in order 
to extract its structural shape features. 
Next, we treat an image which shows a distribution of stars.
In a fifth and last application, we process 1+1D relativistic hydrodynamic 
simulation data in order to obtain a freeze-out hyper-surface for subatomic 
multi-particle production.

\subsection{Region-enclosing contours for EBSD imagery}

\begin{figure}[b]
\epsfig{width=8.5cm,figure=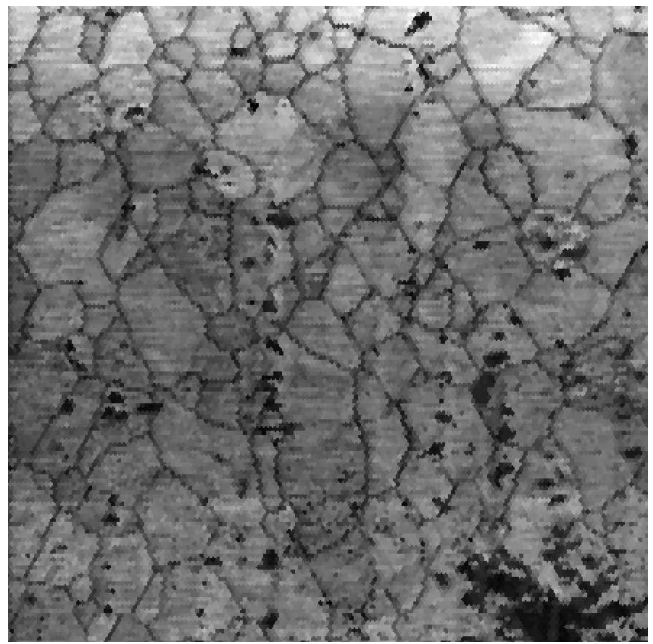}
\vspace{-0.1in} \caption{
Backscattered contrast image~\cite{REF27} of a thin Aluminum film.
} \label{multib_07}
\end{figure}

The accurate characterization of the structures and properties of grain 
boundary networks is one of the fundamental problems in interface science. 
The Electron BackScattered Diffraction~\cite{REF25} (EBSD) technique 
provides experimental results on grain boundary properties and grain 
growth in metal surfaces. In EBSD experiments, images of various material 
surfaces are recorded by secondary electron or backscattered contrast and 
corrected for instrumental distortions. For the extraction of the information
contained in the images, it is important to locate the grain boundaries 
and triple junctions between grains. This localization task is 
accomplished by shape processing techniques, which have been presented 
in the previous sections. The resulting region-enclosing contour 
information is essential for mesh generation~\cite{REF26} and the 
characterization of the morphology and topology of grain distributions.\\
\indent
In this section, we follow the processing of an experimental EBSD image 
with the above decribed shape processing algorithms. Fig.~7 shows an 
example of a backscattered contrast image~\cite{REF27} of a thin 
Aluminum film with a columnar grain structure. Using an image which is 
recorded simultaneously from secondary electron emission, researchers are 
able to determine the grain edge pixels with rather 
standard~\cite{REF08}--\cite{REF10}
pixel processing techniques (\textit{cf.}, Ref.s~\cite{REF25,REF26}; 
\textit{e.g.}, in Ref.~\cite{REF28}, grain boundary information is 
obtained from imagery taken by transmission electron microscopy).\\
\indent
\begin{figure}[t]
\epsfig{width=8.5cm,figure=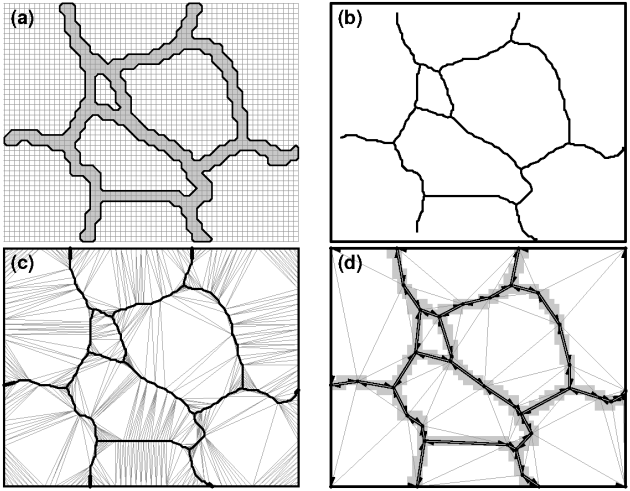}
\vspace{-0.1in} \caption{
Processing a sub-region of the resulting binary image: 
(a) dilated contours enclose the gray pixels; (b) shape skeleton and 
added outer frame; (c) CDT for the final closed contours (the closed 
gaps are drawn with larger line width); (d) gray pixels superimposed 
with down-sampled shape-enclosing contours and a CDT grid.
} \label{multib_08}
\end{figure}
\begin{figure}[b]
\epsfig{width=8.5cm,figure=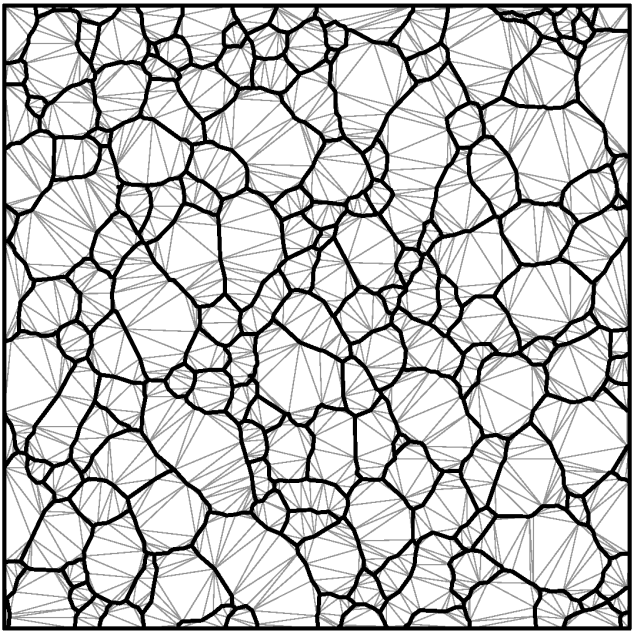}
\vspace{-0.1in} \caption{
Shape-enclosing contours and a CDT grid for the
binarized image of Fig. 7.
} \label{multib_09}
\end{figure}
In the following, we shall restrict our discussion to a sub-region 
(\textit{cf.}, Fig.~8.a) of the resulting bi-level image of grain edge pixels
without loss of generality. The goal is to process the given grain edge pixels 
in order to obtain as a final result a set of contours, where 
\textit{(i)} each contour encloses a grain (region) counter-clockwise with 
a minimum number of supporting points according to the initially given 
edge pixels, and where \textit{(ii)} the whole area of the image has been 
taken under consideration (\textit{cf.}, Fig.~8.d). This can be 
accomplished as follows.\\
\indent
First, right-turning, i.e., pixel connecting, DICONEX contours are 
generated for the gray pixels shown in Fig.~8.a.
Then, a CDT is applied to the dilated contours and their 
supporting point set in order to decompose the interior of the shape into 
triangles. Using the individual morphological roles of the triangles, a 
pruned skeleton (with $\rho_0 = 0.6$) is generated (\textit{cf.}, Fig.~8.b).
It encloses only three of the nine visible grains fully. 
Because it is our intent to construct region-enclosing contours for all 
nine visible grains, a frame (which also is shown in Fig.~8.b) is added 
around the original image area. 
A subsequent CDT is applied to the point set of the skeleton and of 
the image frame (\textit{cf.}, Fig.~8.c). For this current application,
we choose to preserve the orientation of the last line segment for a 
limb-like skeleton arc as much as possible when connecting it to the 
frame (\textit{cf.} Fig.~8.c). Considering the vector sets of (i) the 
shape skeleton, (ii) the image frame, and (iii) the gap vectors, we apply 
the above described contour element connection algorithm (\textit{cf.}, 
section 2.1) in order to connect all vectors into discrete 
region-enclosing contours (note, that only left-turns will be performed, 
whenever contour junctions are encountered). However, the region-enclosing 
contours are rather densely sampled. In Fig.~8.c, the 
contours are sampled with 599 points, and the Delaunay mesh shown here 
consists of 986 triangles. Therefore, we simplify the contours with
the technique explained in Ref.~\cite{REF17} (in particular, we have chosen 
$w_0 = 0.7$ of the pixel width, \textit{cf.}, Ref.~\cite{REF17}). 
In Fig.~8.d, the final contours are sampled with only 38 points, and the 
Delaunay mesh shown here consists of only 64 triangles. Note, that the 
region-enclosing contours stay within the limits defined by the gray 
pixels.\\
\indent
Full processing~\cite{REF17} of the full binary image (not shown here) 
with 25,518 grain edge pixels leads to only 1,368 region-enclosing contour 
support points for 177 grain regions, and the accurate coverage of all 
grains requires only 2,678 Delaunay triangles (\textit{cf.}, Fig.~9).

\subsection{Bacterial colony counting}

\begin{figure}[b]
\epsfig{width=8.5cm,figure=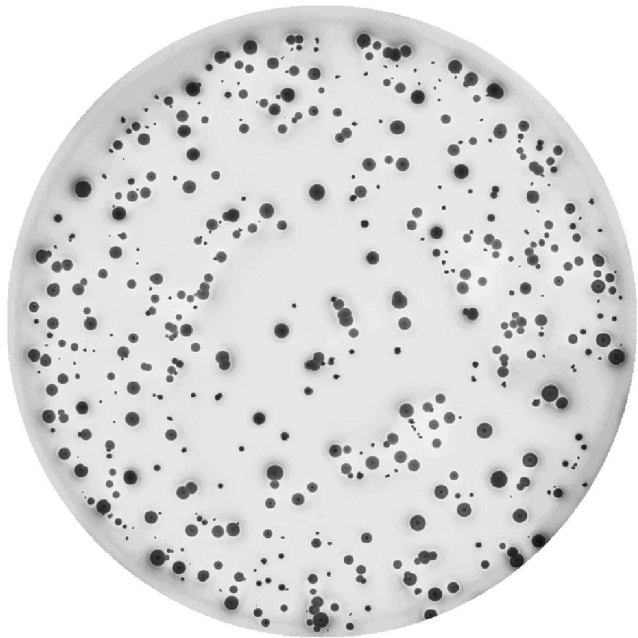}
\vspace{-0.1in} \caption{
Gray-level image~\cite{REF30} of the interior of a 
Petri dish with E. coli colonies.
} \label{multib_10}
\end{figure}

In studies of the population dynamics of the intestinal population of 
mice, the most basic measure of how the population of a certain 
species is behaving is the abundance of the organisms~\cite{REF29}. 
As such, fecal sampling and plating at several dilutions on selective 
media allows the biologist to get a measure of the abundance of organisms 
in the lower colon of mice. Counting colony forming units (CFUs) is a 
traditional way of measuring the population density of any bacterial 
culture or natural bacterial source. Fig.~10 shows the interior of a 
Petri dish with E. coli colonies, which have been recovered from the fecal 
pellets of mice~\cite{REF30}. E. coli is the abbreviated name of the 
bacterium in the family Enterobacteriaceae named Escherichia coli.\\
\indent
In clinical studies, researchers usually take repeated samples for 
statistical reasons. Giving the mice under consideration various 
treatments which alter their intestinal flora results in different 
effects in the population densities. Even a small experiment with -- 
let's say -- 36 mice, will produce hundreds of plates per sample point 
and can be sampled several times daily. The abundance of visual data makes 
the automation of visual bacterial colony counting highly desirable.
However, the separation of overlapping colonies is a challenging task
for standard image processing techniques, i.e., it is not trivial to 
provide the correct number of CFUs present in a given image.
In total one obtains 415 isocontours (for an arbitrarily chosen 
isovalue of 130) for Fig.~10 when applying the contour extraction 
techniques outlined in sections~2.1 and 2.3, respectively. However, 
some of the contours then enclose more than a single CFU. Therefore, this 
number of 415 contours does not reflect the correct number of CFUs present 
in Fig.~10.\\
\indent
In section~2.6, it was explained how one can refine the contours, in order 
to decompose the enclosed shapes into a maximum
of convex shaped constituents (\textit{cf.}, also Ref.~\cite{REF21}). 
In fact, Fig.~6.a is a subregion of the image in Fig.~10 at an increased 
resolution. In section~2.6, we were able to count all of the shown CFUs 
correctly. Fig.~11 shows the refined contours. Note, that the number of 
contours (and therefore the number of registrated CFUs) increases from 415 
to 451 in Fig.~11. However, there are still quite a number of 
contours, which the reader possibly would have refined as well. The contour 
refinement technique as described in section~2.6 is very sensitive to 
the particular result of the constrained Delaunay tessellation. The CDT in 
return, is very dependent on the quality of the contours. Small variations 
in the contours can be caused, e.g., by noise in the image data. However, 
it is beyond the scope of this paper to address a proper treatment of 
noise (as well as other topics such as proper illumination of the sample, 
etc.) for the image data shown in Fig.~10. Hence, we conclude this 
section with the notion that we have obtained a significant 
improvement in our attempt to count bacterial colonies through our new 
framework for contour extraction.

\begin{figure}[t]
\epsfig{width=8.5cm,figure=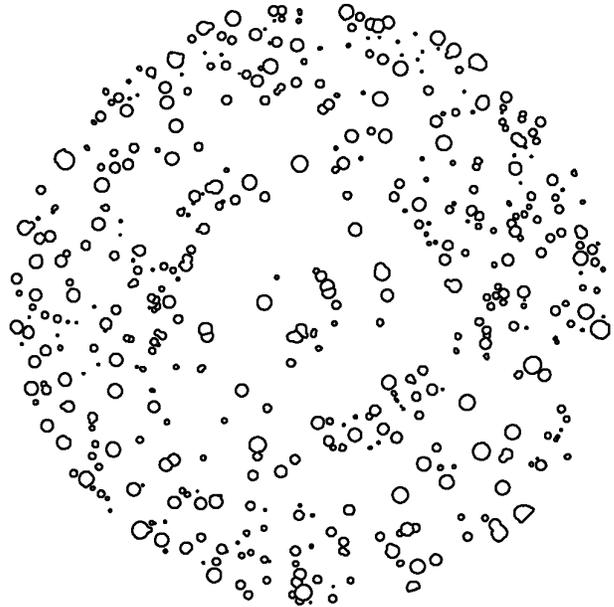}
\vspace{-0.1in} \caption{
Shape-enclosing contours for the E. coli colonies
in Fig. 10. after contour refinement (see text).
} \label{multib_11}
\end{figure}

\subsection{Towards handwritten letter recognition}

Shape analysis and its classification is of great interest within the field
of image processing (\textit{cf.}, e.g., also Ref.~\cite{REF31} and Refs. 
therein).
In section~2.4, a technique has been described which provides skeletons
for contours that enclose pixel regions representing a certain shape.
In fact, a shape skeleton can be viewed as a graphical 
representation~\cite{REF32,REF33} of such an initially given shape 
(\textit{cf.}, also section~2.6).
Refs.~\cite{REF34,REF35} decribe, how a skeleton may be processed 
further so that more knowledge about its represented shape can be gained. 
In particular, such an approach allows for shape recognition. Its detailed
description, however, is beyond the scope of this paper. 
Instead, we provide an example, which demonstrates the transformation of 
the image of a handwritten character into its graphical, i.e., skeletal 
representation.\\
\indent
In Fig. 12.a, a handwritten character, the Japanese Hiragana ``A,'' is 
depicted. All pixels which have a gray-level in the range from $0$ to 
$150$ have been selected and their union has been enclosed with dilated 
contours (\textit{cf.}, Fig.~12.b).
In Fig.~12.c, a CDT has been applied to the point set which supports the
contours. Here, we show both the interior and exterior triangles of the 
mesh. The distinction between interior and exterior triangles is
rather straigthforward, since the dilated contours enclose shapes always
counterclockwise, whereas holes of shapes are always enclosed clockwise.\\
\begin{figure}[t]
\epsfig{width=8.5cm,figure=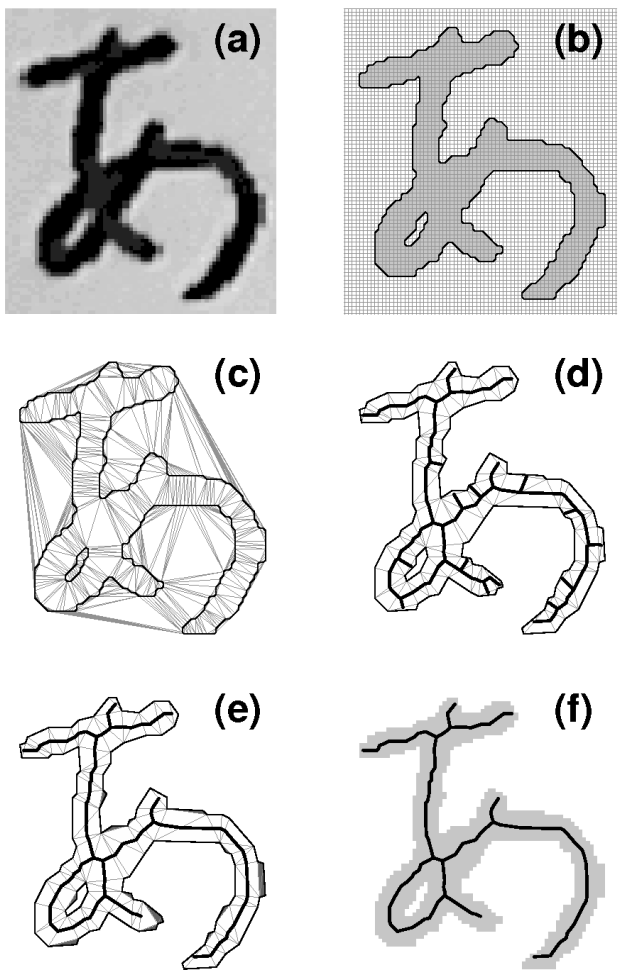}
\vspace{-0.1in} \caption{
(a) A gray-level image resulting from the scan of 
a handwritten Japanese letter, the hiragana ``A;'' (b) dilated contours
enclose the selected gray pixels; (c) CDT for the dilated contours;
(d) CDT and skeleton segments for the interior of the shape, which is
enclosed by down-sampled dilated contours; (e) as in (d), but with a 
pruned skeleton; (f) selected gray pixels superimposed with the pruned
skeleton as shown in (e).
} \label{multib_12}
\end{figure}
\indent
The here shown triangular CDT mesh consists of $526$ interior triangles;
$93$ of them are junction triangles. This rather large number of junction
triangles is mainly caused by the many directional changes in the contours.
After down-sampling of the contours with the technique outlined in 
Ref.~\cite{REF17} (again, we have chosen $w_0 = 0.7$ of the pixel width 
here), we obtain a much coarser CDT mesh. The result is shown in Fig. 12.d, 
where we have only $103$ interior triangles; this time, only $18$ of them 
are junction triangles. However, the shape's skeleton still contains limbs 
which are of apparent lesser morphological significance.\\
\indent
Pruning of the shape with $\rho_0 = 0.4$ yields a skeleton which resembles
its morphology quite well (\textit{cf.}, Fig.~12.e). The final number
of junction triangles is $6$ out of $88$ unpruned triangles.
In Fig. 12.f, the final shape skeleton is superimposed to the original
selected pixels. Note, that the skeleton stays completely within the area
defined by the pixels, which represent the handwritten character. We
conclude this section by refering the more interested reader once again to 
Refs.~\cite{REF34,REF35}, where the subsequent shape recognition processing 
steps are explained in great detail (for alternate pattern recognition
methods,\textit{cf.}, e.g., Ref.~\cite{REF36} and Refs. therein). 

\subsection{Stargazing}

\begin{figure}[b]
\epsfig{width=8.5cm,figure=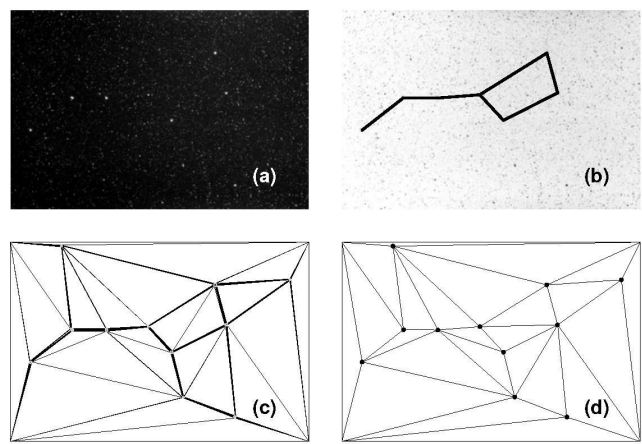}
\vspace{-0.1in} \caption{
(a) A night sky scene of the northern hemisphere
showing the Big Dipper (see text); (b) inverted gray-level image (a) 
with line segments connecting the stars of the Big Dipper; (c) CDT for 
the dilated contours of the pixels representing some of the brigthest 
stars plus the four corners of image (a); (d) CDT for the centers of 
gravity for the star shapes plus the four corners of image (a).
} \label{multib_13}
\end{figure}

For ages, particularly bright stars of the starry sky have been combined 
into constellations~\cite{REF37}. In the northern hemisphere, e.g., the Big 
Dipper -- a group of seven bright stars (\textit{cf.}, Fig.~13.a) that 
dominate the constellation Ursa Major the Great Bear -- can be observed in 
the night sky during any season. Apparently, it is the human eye which 
sometimes can makeup invisible lines, where there are actually 
none~\cite{REF38}. That is why one tends to connect close stars with 
virtual lines, as it is depicted in Fig.~13.b.\\
\indent
In two dimensions, Delaunay tessellations~\cite{REF13} are known to connect 
closer points more likely than points which lie farther apart. In the
following, we are going to process the pixels representing some of the
brightest stars in Fig.~13.a with a CDT, while wondering whether we will be 
able to find all line segments which form the stick figure representation 
of the Big Dipper (\textit{cf.}, Fig.~13.b). Note, that the brightest stars 
apparently also have a larger diameter than the ones which shine more 
weakly.\\
\indent
First, we select all pixels which have a gray-level in the range from $105$ 
to $255$ for the brigtest stars. Next, a dilated contour extraction 
(\textit{cf.}, section~2.1) is performed. In particular, here only contours
above eight times of the length of a pixel are selected, in order to 
account for the brightest stars only. The constructed contours are shown in
Fig.~13.c. After addition of the four image corners to the contour point set, 
a CDT has been performed. Like in Fig.~12.c of the previous 
section~3.3, both the interior and external mesh triangles are shown.
Note, than some mesh edges appear to be thicker than other ones. In fact,
the thicker lines are just very many CDT mesh edges, which almost lie in 
parallel, whereas thinner mesh edges are fewer or just single mesh edges.\\
\indent
In order to find single representative mesh edges for the many which 
almost lie in parallel, we are going to replace the region-enclosing 
contours by centers of gravity~\cite{REF39} of the enclosed triangulated 
areas. Indeed, the contour-enclosed pixels of the stars look very similar 
to the single, cirlular shaped CFUs (\textit{cf.}, also section~3.2) in 
Fig.~6.a. The centers of gravity of the circular shaped areas are calculated 
as follows.
For each triangle in the shape decomposition, we calculate its area and its 
center of mass~\cite{REF39}, respectively. 
Each center of gravity is then the sum of all triangle centers of mass 
weighted by their relative area within a single enclosed shape.\\
\indent
In Fig.~13.d, the centers of gravity are shown as black dots. Once again
a CDT has been performed on the centers of gravity in addition to the
four image corners. As a result, we observe, that all edges of the
stick figure representation of the Big Dipper (\textit{cf.}, Fig.~13.b) are
present in this Delaunay tessellation. In order to deeper understand our 
present findings, one clearly has to gain insights from the field of
human perception (\textit{cf.}, e.g., Ref.~\cite{REF40} and Refs. therein),
which is definitively beyond the scope of this paper. However, we conclude 
this section with the notion that the here presented framework for 2D 
shape-enclosing contours is useful also, when centers of gravity of 2D
shapes need to be calculated.

\subsection{Freeze-out hyper-surface extraction}

Relativistic fluid dynamical models are widely used to describe heavy ion 
collisions~\cite{REF41}. Their advantage is that one can choose explicitly 
the equation of state of the nuclear matter and test its consequences on 
the reaction dynamics and the outcome. This makes fluid dynamical models a 
very powerful tool to study possible phase transitions in heavy ion 
collisions such as the liquid-gas or the quark-gluon plasma phase 
transition~\cite{REF42}. The initial and final, freeze-out stages of the 
reaction are outside the domain of applicability of the fluid dynamical 
model. For example, fluid dynamics is not valid when the fluid becomes 
diffuse. When it is believed that the transition from a fluid to 
subatomic particles occurs (freeze-out), a popular approach for the 
calculation of multi-particle production probability distribution 
functions of hadrons (i.e., subatomic particles, which are composed of 
quarks and qluons) is represented by the integration of source or emission 
functions. The source functions are expressed in the case of relativistic 
hydrodynamic models in terms of hydrodynamic fields ~\cite{REF43} across a 
freeze-out hyper-surface (FOHS).\\
\begin{figure}[b]
\epsfig{width=8.5cm,figure=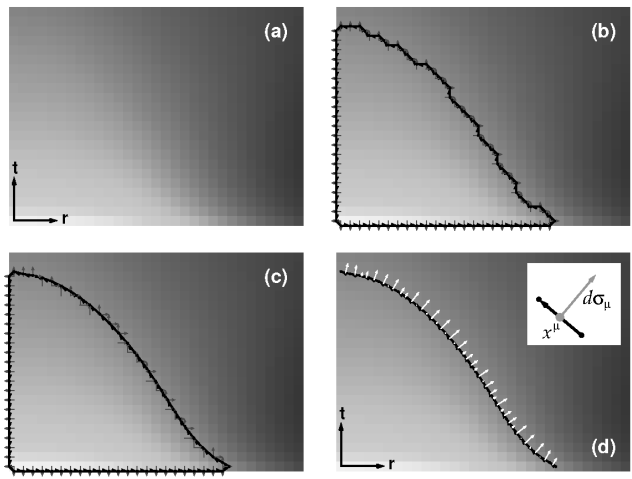}
\vspace{-0.1in} \caption{
(a) A gray-level image representing the temporal 
temperature evolution of a 1D relativistic fluid (see text); 
(b) a dilated contour encloses pixels with temperatures $T \geq T_f$ 
and corresponding range vectors; (c) as in (b), but after contour 
displacement with respect to the fluid temperatures; (d) final FOHS 
with normal vectors $d\sigma_\mu(x^\mu)$.
} \label{multib_14}
\end{figure}
\indent
A FOHS is considered to be a lower-dimensional interface representing the 
union of all subatomic particle production events. In 3+1D (i.e., 3D space 
plus 1D time) hydrodynamic simulations, a FOHS is a 3D volume which is 
embedded in the 4D space-time. Events of subatomic particle production 
take place at a space-time $4$-vector, $x^\mu$ ($\mu = 0,1,2,3$), on the 
FOHS. The index $\mu = 0$ typically refers to the temporal dimension and 
the indices $\mu = 1,2,3$ refer to the three spatial dimensions, 
respectively. 
If one intends to calculate probability distribution functions for the 
production of hadrons, one has to know further 
quantities~\cite{REF44,REF45}. These are, e.g., 
the $4$-normal vector of the FOHS, $d\sigma_\mu(x^\mu)$, the 
$4$-velocity vector of the fluid at freeze-out, $u_\mu(x^\mu)$, the 
temperature at freeze-out, $T_f(x^\mu)$, etc. Very often, the FOHS is 
assumed to be an hyper-isosurface with respect to the temperature 
field~\cite{REF43}--\cite{REF45} of the relativistic fluid, i.e., 
$T_f(x^\mu) = T_f = const$. 
The exploitation of spatial symmetries may allow the physicist to investigate 
certain aspects of relativistic fluid dynamics simulations in reduced 
dimensions, such as in a 1D radial space plus 1D time. Then the problem of 
FOHS extraction becomes identical to a 1D thermal isocontour extraction on 
a discretized 2D hydrodynamic simulation history. The 2D hydrodynamic 
simulation history is comprized of a discretized 1+1D space-time lattice 
(similar to 2D image data), on which field quantities such as temperature 
or fluid velocity components, etc., have been stored.\\
\indent
Fig.~14.a shows a 2D hydrodynamic simulation history of the discrete
fluids temperature field, i.e., the temporal temperature evolution of a
1D relativistic fluid. In other words, Fig.~14.a is an image, $T = T(t,r)$,
where $T$, $t$, and $r$ are the (continuous) fluid temperature, the 
(discretized) time and a (discretized) spatial dimension (e.g., radius, 
because a radial symmetry may apply), respectively. 
Darker pixels refer to lower fluid temperatures, whereas brighter ones 
refer to higher fluid temperatures. The origin of the space-time lattice 
(i.e., $r = t = 0$) is located in the center of the lower left image 
pixel. The FOHS is defined here as a thermal isocontour of value $T_f$ 
and is constructed as follows. First, all lattice points (i.e., 
pixels) which have a temperature higher or equal to $T_f$ are enclosed 
with a left-turning (i.e., pixel disconnecting) DICONEX contour. This is 
depicted in Fig.~14.b, which also shows the range vectors neccessary for 
the following isocontour extraction. In the next step, the isocontour 
is constructed (\textit{cf.}, Fig.~14.c). When the contour support points
are relocated through linear interpolation, the corresponding
field quantities such as fluid velocity field components, etc., are 
evaluated for the isocontour supporting points as well. Note, that points 
which sit directly on the boundary of the image data are not moved. 
Contour edges which have both supporting points with coordinates 
$r \leq 0$ and/or $t \leq 0$, are physically irrelevant (for reasons, 
which are not explained here). They have been removed in the final 
result shown in Fig.~14.d. 
In Fig.~14.d, we also show the corresponding $4$-normal vectors of the 
FOHS at freeze-out, $d\sigma_\mu(x^\mu)$. Note, that in 1+1D these 
$4$-vectors are just the normal vectors of the isocontour 
vectors~\cite{REF46}. 
Furthermore, the 1+1D freeze-out events, $x^\mu = (t_f,r_f)$, are 
associated with the middle of each isocontour vector (\textit{cf.}, 
Fig.~14.d), and so are all corresponding field quantities (i.e., 
$d\sigma_\mu(x^\mu)$, $u_\mu(x^\mu)$, etc.). $t_f$ and $r_f$ denote the 
freeze-out times and freeze-out radii, respectively.

\section{Summary}

In summary, we have introduced a new framework for 1D contour extraction
from discrete 2D data sets. Within this toolbox approach, we can generate
up to five different types of contours. These are (i) the contours made
up by the connected sets of contour vectors which initially separate pairs
of pixels, (ii) DICONEX or dilated contours, (iii) boundary pixel tracing
contours, (iv) isocontours, and (v) contours from shape skeletons, 
respectively. All of the contours can be computed rather fast and 100\%
robustly.
In particular, the DICONEX contours resemble a class of perfect contours
in the sense that they are always non-selfintersecting and non-degenerate, 
i.e., they always enclose an area larger than zero.\\
\indent
An important integral part of the contour extraction toolbox is a 
constrained Delaunay tessellation tool, which aids the gap closure and/or 
continuation of contour fragments such that closed contours may be obtained 
at all times.
The Delaunay tessellations are also useful for contour simplification 
(\textit{cf.}, Ref.~\cite{REF17}), as well as extraction and potential 
pruning of shape skeletons.
The introduction of high-level shape constituents (e.g., torsos) allow
for contour refinement through 2D shape manipulations.\\
\indent
Finally, we have demonstrated that a wide range of rather diverse 
applications can be addressed with this novel contour extraction framework. 

\section{Acknowledgements}

This work has been supported by the Department of Energy under contract
W-7405-ENG-36.\\

\end{document}